\documentclass{article}

\usepackage[preprint]{neurips_2020}

\usepackage[english]{babel}
\usepackage[utf8x]{inputenc}
\usepackage{amsmath}
\usepackage{graphicx}
\usepackage[colorinlistoftodos]{todonotes}
\usepackage{caption}
\usepackage{graphicx}
\usepackage[caption=false]{subfig}
\usepackage{booktabs}

\usepackage[T1]{fontenc}

\usepackage{geometry}
\usepackage{amssymb}

\usepackage{adjustbox}
\usepackage{soul}
\usepackage{multicol}
\usepackage{calc}
\usepackage{floatrow}
\usepackage{listings}
\usepackage{gensymb}  
\usepackage{rotating}
\usepackage{hhline}
\usepackage{multirow}
\usepackage{wrapfig}
\usepackage{hyperref}
\usepackage[ruled,vlined]{algorithm2e}

\setcitestyle{numbers,square}
\title{Causal Future Prediction in a Minkowski Space-Time}

\author{
	Athanasios Vlontzos~\footnote{Corresponding Author} \\
	Imperial College London \\
	London, UK\\
	av2514@ic.ac.uk
	\And
	Henrique Bergallo Rocha \\
	The University of Edinburgh \\
	Edinburgh, UK\\
	h.b.rocha@ed.ac.uk
	\And 
	Daniel Rueckert\\ Imperial College London \\
	London, UK\\
	d.rueckert@ic.ac.uk
	\And 
	Bernhard Kainz\\ Imperial College London \\
	London, UK\\
	b.kainz@ic.ac.uk
}	

\begin{document}
	
	\maketitle
	
	\begin{abstract}

		Estimating future events is a difficult task. Unlike humans, machine learning approaches are not regularized by a natural understanding of physics. In the wild, a plausible succession of events is governed by the rules of causality, which cannot easily be derived from a finite training set. In this paper we propose a novel theoretical framework to perform causal future prediction by embedding spatio-temporal information on a Minkowski space-time. We utilize the concept of a light cone from special relativity to restrict and traverse the latent space of an arbitrary model. We demonstrate successful applications in causal image synthesis and future video frame prediction on a dataset of images. Our framework is architecture- and task-independent and comes with strong theoretical guarantees of causal capabilities.
	\end{abstract}
	
	\section{Introduction}
    In many everyday scenarios we make causal predictions to assess how situations might evolve based on our observations and experiences. Machine learning has not been developed to this level yet, though, automated, causally plausible predictions are highly desired for critical applications like medical treatment planning, autonomous vehicles and security.     
   % A system able to automate the task of causal prediction would be able to assist doctors in treatment planning, enable defensive driving capabilities on autonomous vehicles or support counterfactual analysis. 
    Recent works have contributed machine learning algorithms for the prediction of the future in sequences and for causal inference~\cite{Kurutach2018}.
	One major assumption that many approaches implicitly adopt, is that the space of the model representation is a flat Euclidean space of N dimensions. However, as shown by Arvanitidis et al.~\cite{arvanitidis2018latent}, the Euclidean assumption leads to false conclusions as a model's latent space can be better characterized as a high dimensional curved Riemannian manifold rather than an Euclidean space. Furthermore, the Alexandrov-Zeeman theorem~\cite{Zeeman1964,kosheleva2014observable} suggests that causality requires a Lorentzian group space and advocates the unsuitability of Euclidean spaces for causal analysis. 
	%Furthermore, the choice of the representation space affects more than the way the latent codes are interpreted. Appropriate changes have to be adopted in both the architecture of the networks as well as the optimization in order to make them compatible with Riemannian spaces. For example gradient descent has to be substituted by Riemannian gradient descent that incorporates the manifold's metric~\cite{NIPS2015_5971}.
	
	In this paper, we present a novel framework that changes the way we treat hard computer vision problems like the continuation of frame sequences. We embed information on a spatio-temporal, high dimensional pseudo-Riemannian manifold - the Minkowski space-time - and utilize the special relativity concept of light cones to perform causal inference. We focus on temporal sequences and image synthesis to exhibit the full capabilities of our framework.
	
	In summary our contributions are: 
	\begin{itemize}
	    \item We extend representation learning to spatio-temporal Riemannian manifolds that follow the ideas of the Minkowski space-time while being agnostic towards the used embedding architecture and the prescribed task.
	    \item We introduce a novel utilization of the concept of light cones and use them for convincing frame synthesis and plausible prediction of future frames in video sequences. 
	    \item We provide theoretical guarantees about the causal properties of our model and demonstrate a causal inference framework.
	\end{itemize}
	
\section{Related Works}
%The use of  in machine learning literature can be found in a few major works. 
High dimensional Riemannian manifolds for machine learning are utilized by  a few major works.
Arvanitidis \emph{et al.}~\cite{arvanitidis2018latent}  show evidence that more general Riemannian manifolds characterize learned latent spaces better than an Euclidean space. Their work however, utilizes generators that have been trained under an  Euclidean assumption. 
Contrary to that, Nickel \emph{et al.}~\cite{Nickel2017} introduce the use of a Poincar\'e ball for hierarchical representation learning on word embeddings, showing superior performance in representation capacity and generalization ability while employing a Riemannian optimization process. In \cite{Nickel2018}, Nickel \emph{et al.} extend the previous work to a Lorentzian manifold as this offers improvements in efficiency and stability of the distance function. In this paper we accept the argument made by Nickel \emph{et al.} but extend it as we argue in Section~\ref{why?} that causal inference requires a Lorentzian group space~\cite{Zeeman1964}. 

Ganea~\emph{et al.}~\cite{Ganea2018} embed word information on a Poincar\'e ball and form entailment cones. The authors propose to work with Directed Acyclical Graphs (DAG) and strive for non overlapping cones in a Poincar\'e ball. In contrast to this,  we encourage overlapping light cones in a Lorentzian manifold to model future events. 

Sun \emph{et al.}~\cite{NIPS2015_5971} use a space-time idea similar to ours but interpret the time axis as a ranking rather than as temporal information. Their method is intended for dimensionality reduction and does not generate further samples, or considers causal relationships between sampling points. Finally, Mathieu \emph{et al.}~\cite{Mathieu2019} train a Variational Autoencoder (VAE) constrained to a Poincar\'e ball while also employing the appropriate Riemannian equivalent to a normal distribution as well as Riemannian optimization. 
We consider this work as the closest related since it is the only approach that has shown good performance in the image domain.

In the Computer Vision focused field of future frame prediction for video sequences, \cite{Kurutach2018} propose the causal InfoGAN which, however, lacks theoretical guarantees of causal abilities.
\cite{jayaraman2018timeagnostic} aims at predicting the probabilistic bottlenecks where the possible futures are constrained instead of generating a single future. Similarly, we are not attempting to predict a single future, rather we  predict all plausible futures in a way that naturally enables us to  identify all probabilistic bottlenecks; see Section~\ref{entropy}. In other works concerned with video continuation, \cite{MathieuCL15, cvpr/VondrickPT16} use CNNs to regress future frames directly, while~\cite{VillegasYHLL17} introduce an LSTM utilizing the difference $\Delta$ between frames to predict motion. Further works include the use of optical flow~\cite{LiuLLG18} or human pose priors~\cite{DBLP:journals/corr/VillegasYZSLL17}. The autoregressive nature of these methods results in accumulated prediction errors that are propagated through the frames the further a sequence is extended. Beyond a few frames, these approaches quickly lose frame-to-frame consistency. In order to mitigate these limitations, works like \cite{nips/VondrickPT16} propose generative models to predict future frames and \cite{Tulyakov0YK18} offers a generative model that disentangles motion and content. Neither can infer the causal implications of their starting positions. % Finally the aforementioned models have no guarantees that their resulting predictions can be causal consequence of their starting position.%Finally, \cite{BabaeizadehFECL18} propose a video prediction model based on probabilistic graphical model.

\section{Theoretical Formulation}
\subsection{Background}
For an in-depth review of manifolds we invite the reader to refer to S. Carroll's textbook \cite{Carroll1997}; in this section we will briefly provide some key concepts of the used differential geometry. 

\textbf{Manifold:} A manifold $M$ of dimensions $n$ is a generalization of the concept of the surface in a non-Euclidean space and is characterised by a curvature $c$. The manifold group that this paper considers is constantly flat.

\textbf{Tangent Space:} The tangent space $T_xM$ is a vector space that approximates the manifold $M$ at a first order.

\textbf{Riemannian Metric:} A Riemannian metric $g$ is a collection of inner products $T_xM \times T_xM \rightarrow {\Bbb R}$. It can be used to define a global distance function as the greatest local bound of the length $l$ of all the smooth curves $\gamma$ connecting points $x,y \in M$. Note that the length is defined as $l(\gamma) =\int _{0}^{1}\sqrt{\mathrm{g_{\gamma(t)}(\gamma'(t),\gamma'(t))}}\, dt $. 

\textbf{Geodesic:} Geodesics are the generalizations of straight lines in Euclidean space and define the shortest path between two points of the manifold. They can also be defined as curves of constant speed.

\textbf{Exponential map:} The exponential map $exp_x:T_xM\rightarrow M$ around $x$ defines the mapping of a small perturbation $v\in T_xM$ to a point in $M$ s.t. $t\in [0,1] \rightarrow exp_x(tv)$, which is the geodesic of $x$ to $exp_x(v)$.

\subsection{Causal Inference}
Causal inference refers to the investigation of causal relations between data. There is a rich literature on machine learning and causal inference ranging from association of events to counterfactuals \cite{Peters2019,Pearl2019}. Briefly we observe two equivalent approaches towards causality in machine learning: Structural Causal Models~\cite{Pearl2019,pearl_glymour_jewell_2016} which rely on Directed Acyclical Graphs (DAG) and Rubin Causal Models~\cite{Rubin2005} which rely upon the potential outcomes framework. In this paper we will be focusing on the latter. In the potential outcomes framework as established by~\cite{Rubin2005} multiple outcomes $\mathcal{Y}$ of $\mathcal{X}$ are contrasted in order to deduce causal relations between $\mathcal{Y}$ and $\mathcal{X}$. As we will show, our proposed method provides the theoretically guaranteed infrastructure to create a Rubin Causal Model. In addition, as our method is able to operate in a future as well as a past regime it enables the formation of counterfactual questions,  \emph{i.e.}, what would $\mathcal{Y}$ be if $\mathcal{X'}$ had happened instead.

\subsection{On the choice of space\label{why?}}
In his seminal 1964 work, E.C. Zeeman~\cite{Zeeman1964} makes the case that the causality group $\mathcal{R}M$ that arises from the concept of partial ordering in a Minkowski space-time implies an inhomogenous Lorentz group as the symmetry group of $\mathcal{R}M$. We highlight the explicit mention of Zeeman on the unsuitability of an Euclidean topology to describe $\mathcal{R}M$ due to its local homogeneity, which does not arise in $\mathcal{R}M$. In ~\cite{kosheleva2014observable} the authors prove that from observable causality we can reconstruct the Minkowski space-time. Hence, we are in a position to argue that the use of a Minkowski space-time for embeddings, which belongs to the inhomogenous Lorentz group, would reinforce causal inference capabilities. 

We define our Minkowski space-time to be characterized by the metric of Eq.~\ref{eq_metric} with the element $-1$ denoting the temporal dimension and $+1$ elements the spatial dimensions. We extend~\cite{Nickel2018} and argue that the use of the Lorentzian manifold, which coincides with the Minkowski space-time, is both more efficient as an embedding as well as enabling causal arguments, 
\begin{equation} \label{eq_metric}
    \eta_{\mu\nu}=\text{diag}(-1,+1,+1,+1).
    \end{equation}

\subsection{Minkowski Space-Time and Causality}
Mathematically a space can be described by its metric, which defines the way the inner product of two vectors in this space is determined, \emph{i.e.} the way we calculate distances. % In this paper we use the Minkowski space-time whose metric tensor $\eta_{\mu\nu}$ is formulated in Eq~\ref{eq_metric}.
Consequently, the inner product $\langle.,.\rangle_\eta$ of two vectors $a$ and $b$ in $1+3D$ Minkowski space-time can be defined as
  \begin{equation} \label{Minkowski}
    \langle a,b\rangle_\eta=\sum_{\mu=0}^3\sum_{\nu=0}^3 a_\mu \eta_{\mu\nu}b_\nu=-a_0b_0+a_1b_1+a_2b_2+a_3b_3,
  \end{equation}
  where the coordinate 0 is understood to be the time coordinate.

One of the consequences of endowing the latent space with a Minkowski-like metric is the emergence of causality in the system. This property can be more readily seen by employing the concept of \textit{proper time}. Given a manifold $\mathcal{M}$ endowed with a Minkowski metric $\eta_{\mu\nu}$, we define the proper time $\tau$. This is the time measured by an observer following along a continuous and differentiable path $\mathcal{C}(s)$ parametrized by $s\in[0,1]$ between two events $\{x,y\} \in \mathcal{M}$ such that $\mathcal{C}(0)=x$, $\mathcal{C}(1)=y$, 
\begin{equation} \label{proptime}
\tau_{\mathcal{C}} = \int_\mathcal{C}\sqrt{-\sum_{\mu, \nu}dx_\mu dx_\nu}.
\end{equation}

In order to ensure $\tau \in \mathbb{R}$, we require $\sum_i dx_i^2\leq dx^2_0$, where $i\in{1,2,...,d}$. Therefore, the rate of change $|\mathbf{dx}|/d\tau$ in the spatial coordinates is capped by the time evolution of the system. In other words, there exists a maximum speed limit which $\mathcal{C}$ must obey at every point. Further, it means that there exist pairs of space-time points ${x,y}$ which cannot be possibly connected by a valid path $\mathcal{C}$ , lest $\tau \notin \mathbb{R}$. In order to describe this phenomenon we borrow the concept of a \textit{light cone} from special relativity. The set of solution paths $\{\mathcal{C}_0(s)\}$ such that $\mathcal{C}_0(0)=(t_0,\mathbf{x_0})$ and $\tau_{\mathcal{C}_0}=0$ describe the fastest any particle or piece of information can travel within the system starting from $(t_0,\mathbf{x_0})$. This boundary is known as the light cone, and is such that $\partial\mathcal{R}=\{\mathcal{C}_0(s)\}$, where $\mathcal{R}$ is the causal region of the point $(t_0,\mathbf{x_0})$. Every space-time point $x\in\mathcal{R}$ is said to be within the light cone. As shown by (\ref{proptime}), no valid path $\mathcal{C}(s)$ can cross $\partial\mathcal{R}$. Thus, two space-time points can only influence each other if they lie within each other's light cone, that is, if they can be connected by a valid path $\mathcal{C}$. The region $\mathcal{R}$ splits into two disjoint sets: $\mathcal{R}^+$ and $\mathcal{R}^-$.  $\mathcal{R}^+$ lies within the future light cone of a particle at time $t_0$, and thus includes all of the points $(t_1,\mathbf{x_1})\in\mathcal{R}$ such that $t_1>t_0$. Conversely, $\mathcal{R}^-$ includes the points $(t_2,\mathbf{x_2})\in\mathcal{R}$ such that $t_2<t_0$ and characterizes the past light cone of a particle at time $t_0$. 

If we have two space-time vectors  $x=(t_0,\mathbf{x_0})$ and $y=(t_1,\mathbf{x_1})$ we can describe their relation as \textit{timelike} when $\langle x,y \rangle<0$, \textit{spacelike} when $\langle x,y \rangle>0$ and  \textit{lightlike} when $\langle x,y \rangle=0$. A timelike position vector lies within the light cone of a particle at the origin of the system. A spacelike vector lies outside of it, and a lightlike vector lies exactly at its edge. One can then generalize this idea beyond the origin, and thus compute the inner product of the difference between two space-time vectors $x-y\equiv(\Delta t, \Delta \mathbf{r})$ , \emph{i.e.}, 
% %what the heck is wrong with typesetting here
%\begin{equation}
$
\langle y-x,y-x \rangle = -\Delta t^2+|\Delta \bold{r}|^2. 
$ 
%\end{equation}
Hence, when the separation of the vectors $x$ and $y$ is timelike, they lie within each other's causal region. In that case we can argue that there is a path for particle $x$, that belongs in the model that defines the latent world of represented data, to evolve into particle $y$ within a time period $\Delta t$. Thus, by constructing the light cone of an initial point $x$ we can constrain the space where the causally resulting points may lie. We can then see that this mathematical construction of the latent space naturally enforces that the velocity of information propagation in the system be finite, and that a particle can only be influenced by events within its past light cone, \emph{i.e.} the model is causal. 
By mapping this into a machine learning perspective we argue that in a latent space that is built to follow the Minkowski space-time metric an encoded point can then be used to create a light cone that constrains where all the causally entailed points may be encoded to or sampled from. 

\subsection{On Intersecting Cones}
\begin{figure*}[htb]
    \centering
    \subfloat[Visualization of the emerging structure of a light cone. The intersecting plane at point $z=3$ signifies the 2-dimensional feature space at time 3. The interior of the cone subspace contains all possible frames given a original video frame at point $z=0$.
	\label{cone1}]{
        \includegraphics[width=.45\linewidth]{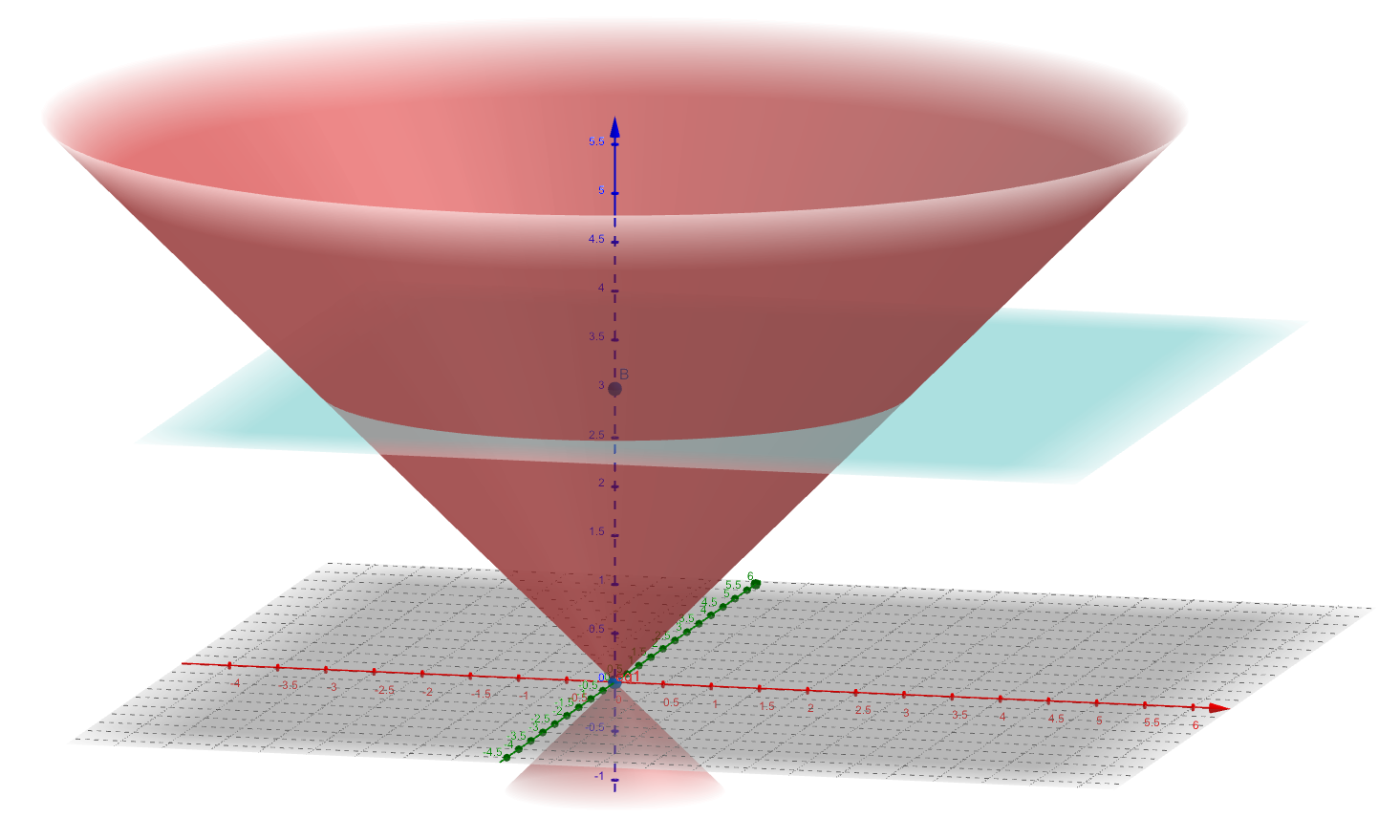}
        }
    \hfill
    \subfloat[Visualization of the intersecting cones algorithm. The subspace marked in yellow contains the points that are causally related to points $F_{0,1,2}$.
	\label{intersection-viz}]{
        \centering
        \includegraphics[width=.4\linewidth]{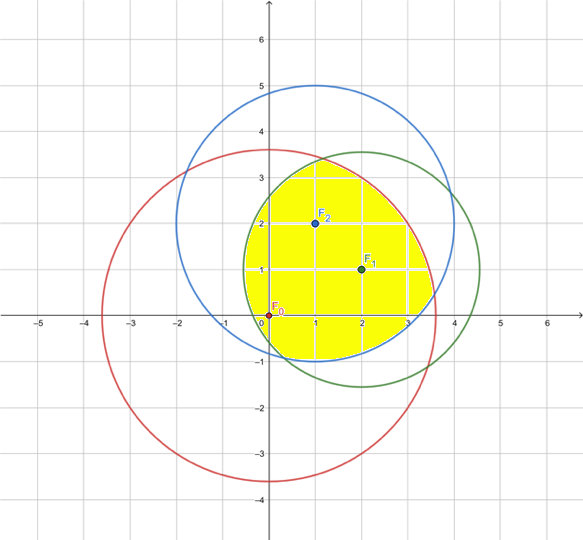}
    }
    \caption{Visual aids of proposed algorithm.  Note that for visualization purposes we are exhibiting a $1+2$ dimensional Euclidean space rather than a high dimensional Riemannian manifold.}
    \label{fig:overview}
\end{figure*}

A light cone can be constructed with each point of the latent space as its origin. 
Consider point $x_0$ to be an initial point derived from,  for example, an encoded frame $f_0$ from a video sequence:  by constructing the light cone $C_0$ around $x_0$ we are able to deduce where the various causally related $x_{0+t}$ points might lie. 
By setting $t$ to be a specific time instance, we are able to further constrain the sub-space to points that lie inside of the conic section. They are causally plausible  results of point $x_0$ within the time $t$. Geometrically,  we can visualize this as a plane cutting a cone at a set time. We visualize this in Figure~\ref{cone1}.    

   % \begin{figure}
%	\includegraphics[width=.5\linewidth]{cone1.png}
	
%	\caption{Visualization of the emerging conical structure for the case $\delta_{0x}=\delta_{0y}$; The intersecting plane at point $z=3$ signifies the 2-dimensional feature space at time 3. The interior of the cone subspace contains all possible frames given a original video frame at point $z=0$. Note that for visualization purposes we are exhibiting a $1+2$ dimensional Euclidean space rather than a high dimensional Riemannian manifold.
%	\label{cone1}}
%	\end{figure}

A second point $x_1$ that lies inside the light cone of $x_0$ can be derived from an encoded frame $f_1$. Similar to $x_0$ we construct the light cone $C_1$ whose origin is $x_1$. We then define the conic intersection $CS = C_0 \cap C_1$. Following the causality argument, we deduce that the enclosed points in $CS$ are causally related to both $x_0$ and $x_1$ as they lie in the light cones of both. In addition, by constraining the intersecting time plane, we constrain the horizon of future prediction.
	
Consequently, we propose Algorithm~\ref{algo1} as a method of future frame prediction using light cones on a Minkowski space-time latent space. We graphically represent Algorithm~\ref{algo1} in Figure~\ref{intersection-viz}.
	
%\begin{figure}
%	\includegraphics[width=.5\linewidth]{top_down_intersection.png}
	
%	\caption{Visualization of the intersecting cones algorithm. The subspace marked in yellow contains the points that are causally related to points $F_{0,1,2}$. Note that for visualization purposes we are exhibiting a $1+2$ dimensional Euclidean space rather than a high dimensional Riemannian manifold.
%	\label{intersection-viz}}
%	\end{figure}
	
	\begin{algorithm}[H]
	\label{algo1}
\SetAlgoLined
\SetKwInOut{Input}{Input}\SetKwInOut{Output}{Output}
\Input{Frame Sequence $F$ ; Queried Time $T$}
\Output{Predicted Frame}

 \For{$t < T$}{
  $Mf_t \leftarrow MinkowskiEmbedding(f_t) $ \\
  $C_{Mf_t} \leftarrow LightCone(Mf_t) $\\
  \If{$t > len(F)$}{
    $Samples_{Mf_t} \leftarrow sample(C_{Mf_t})$ \\  
    $Mf_{t+k} \leftarrow choose(Samples_{Mf_t})$
   }
   }
 $CS \leftarrow intersection (C_{MF})$\\
 $f_{out} \leftarrow choose(sample(CS))$ \\ 
 Predicted Frame $\leftarrow Decoder(f_{out})$
 \caption{Future Prediction using Intersecting Light Cones}
\end{algorithm}
	
\subsection{On the Entropy and the Aperture of Cones \label{entropy}}
When considering the intersection of the cones in Algorithm~\ref{algo1} it is vital to examine the aperture of the cone at time $T$. For simplicity, we assume that the gradient of the side of the cone is $45\degree$ for all cones. However, such an assumption implies that each frame and hence each cone evolves with the same speed and can reach the same number of states at a given time. For real world scenarios this is not necessarily true as, for example, the possible states in $t+1$ for a ball rolling  constraint by rails are less than a ball rolling on a randomly moving surface. Hence, the actual gradient of the cone depends on the number of states that are reachable from the state depicted in frame $t$. This quantity is also known as the thermodynamic entropy of the system. It is defined as the sum of the states the system can evolve to. Calculating the thermodynamic entropy of a macro-world system as in a real world dataset is not trivial and we are not aware of any appropriate method to compute this at the time of writing. Hence, we are forced to make the aforementioned assumption of $45\degree$. 

However, given a frame sequence $F$, a set of counter example frames $CF$ and following Algorithm~\ref{algo1} but omitting the sampling steps, it is possible to build more accurate light cones in a contrastive manner. Hence, it is possible to acquire a proxy for the thermodynamic entropy of the system. We note that the proxy can only be accurate to a certain degree as any frame sequence is not able to contain enough information to characterize the full state of the world. %This also is the reason why the intersecting cones do not overlap in their entirety.

\section{Experimentation}
\subsection{Training}
Our proposed algorithm is invariant to the method used to train the embedding. In an ideal scenario, we require an encoder-decoder pair that is able to map any image to a latent space and to reconstruct any latent code. For the purposes of this paper's evaluation we have chosen the method by Mathieu \emph{et al.}~\cite{Mathieu2019} as our baseline embedding, as it is the only approach that has shown good image domain performance. 

Mathieu~\emph{et al.}~\cite{Mathieu2019} construct a Variational Auto Encoder (VAE) that enforces the latent space to be a Poincar\'e Ball. We analyze the properties of the Poincar\'e ball in the supplementary material. It can be shown~\cite{Nickel2018} that a $n-$dimensional Poincar\'e ball embedding can be mapped into a subspace of the Minkowski space-time by an orthochronous diffeomorphism $m: P^n \rightarrow M^n$,
\begin{equation}
 m(x_1,...x_n) = \frac{(1+||x||^2,2x_1,...,2x_n)}{1-||x||^2},  
  \label{eq:m}
\end{equation}
and back with the inverse $m^{-1}: M^n \rightarrow P^n$
\begin{equation}
 m^{-1}(x_1,...x_n) = \frac{(x_1,...,x_n)}{1+x_0},   
 \label{eq:m-1}
\end{equation}
where $x_i$ is the i-th component of the embedding vector.

We extend~\cite{Mathieu2019} to enforce the embedding to a subspace of the Minkowski space-time by utilizing Eq.~\ref{eq:m} and \ref{eq:m-1}. We treat the space's dimensionality as hyper-parameter and tune it experimentally. We establish that the optimal embedding of our data can be achieved in an $1+8$ dimensional space \emph{i.e.} 1 time and 8 space dimensions. The model consists of a MLP with a single hidden layer and was trained with the Riemannian equivalent of the Adam optimizer~\cite{NIPS2015_5971} with a learning rate of $5e-4$. Training the model with Moving MNIST requires on a Titan RTX Nvidia GPU less than 1 hour. 

\subsection{Inference}
Our proposed Algorithm~\ref{algo1} is executed during inference as it does not require any learned parameters. We sample from a Gaussian distribution wrapped to be consistent with our Minkowski space-time in a manner similar to~\cite{Mathieu2019}, details can be found in the supplement. Inference  can be performed in about $0.5$ s per intersecting cone in an exponential manner.

\subsection{Dataset} As a proof of concept we use  a custom version of the Moving MNIST dataset\cite{DBLP:journals/corr/SrivastavaMS15}. Specifically we employ 10.000 sequences consisting of 30 frames each, making a total of 300.000 frames. Each sequence contains a single digit.  %as opposed to two of the original dataset.
The first frame is derived from the training set of the original MNIST dataset, while the subsequent frames are random continuous translations of the digit. Construction of the test set followed the same procedure with the first frame derived from the test set of the original MNIST dataset. We created 10.000 testing sequences of 30 frames each. Each frame is $32\times32$ while the containing digits range from $18px-25px$

We further use the KTH action recognition dataset~\cite{schuldt2004recognizing} to highlight the real world capabilities of our method. We focus on the walking and handwaving actions and use all 4 distinct directions. Different person identities are used in the train-test split.
\section{Results}

\subsection{Experiment 1: Single Cone Image Synthesis}
In the first experiment we evaluate the ability of the light cone to constrain the latent space such that samples lying inside the cone are reasonably similar to the original frame. We train our model with 1+8 latent dimensions.
Following standard VAE sampling we produce 100.000 random samples using a wrapped normal distribution. As expected, the tighter the imposed time bound is, the fewer samples are accepted. We note that for $t=2$ only $2$ samples were accepted, for $t=10$ our method accepts $N=31\%$ of the samples and for $t=20\text{, } N=71\%$.
In Figure~\ref{exp1} we exhibit qualitative results for Experiment 1. We note that as the time limit increases we observe higher variability, both in terms of morphology and location of the digits, while the identity of the digit remains the same. This is in accordance with the theory that the ``system'' would have enough time to evolve into new states. More examples are included in the supplementary material.  

 \begin{figure*}[htb]
    \centering
    \subfloat[Samples from Experiment 1.
    \label{exp1}]{
        \includegraphics[width=0.5\linewidth]{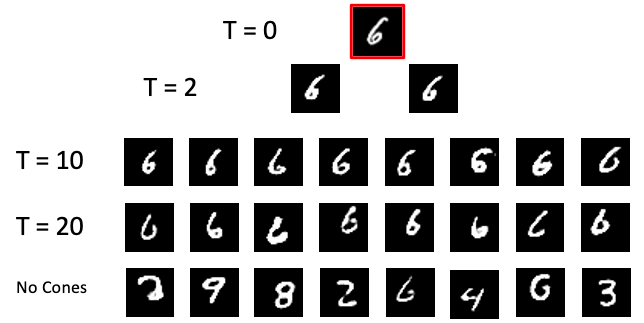}
        }
    \subfloat[Samples from Experiment 2.
    \label{exp2_a}]{
        \includegraphics[width=0.5\linewidth]{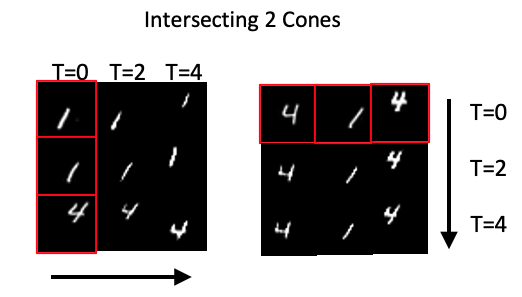}
        }
        \caption{(a): Random sampling was constrained in Experiment 1 such that the samples lie inside the light cone with an upper temporal bound. Samples in the last row of Figure (a) had no constraints imposed on them. We observe larger morphological and location differences as time progresses. This is consistent with the theory that the system had enough time to evolve into these states. (b): In Experiment 2 we are intersecting 2 cones. For ease of reading the figures have been arranged such that the movements are more apparent.  on the left in (b) we exhibit vertical movements while on the right we exhibit horizontal movements. The arrows guide the direction of reading in the figure.}
\end{figure*}

\iffalse
\begin{figure}
	\includegraphics[width=.6\linewidth]{images/exp1/exp1.png}
	\caption{Samples from Experiment 1. Random sampling was constrained such that the samples were in the light cone with an upper temporal bound. Samples in the last row had no constraints imposed on them. We observe larger morphological and location differences as time progresses. This is consistent with the fact that the system had enough time to evolve into these states.
	\label{exp1}}
	\end{figure}
\fi

\subsection{Experiment 2: Intersecting Cones}
In the second experiment we evaluate the ability of our algorithm to predict frames by intersecting light cones.
There is no unique path a system might evolve in time. Our algorithm does not aim at producing a single future, rather it is able to produce multiple plausible futures. At a single time instant we can find any number of probable frames that extend a sequence. Hence, the {\tt{choose}} step of Algorithm \ref{algo1} depends on the target application. In this experiment to guide the choice of frames we map the sampled points to image space and compare the structural similarity of them with the original frame of $t=0$. We adopt a simple manner to choose the next step and we do not provide the model with any further conditioning information to highlight the default strengths of the proposed algorithm. In an online inference scenario the reference frame could be updated as ground truth frames become available. 

In Figures \ref{exp2_a} and \ref{exp2_b} we exhibit qualitative results of our algorithm when intersecting 2 and 5 cones respectively. In Figure~\ref{exp2_a} each set of results evaluates a specific movement, vertical or horizontal. In Figure~\ref{exp2_b} we exhibit the case of intersecting 5 cones. As this scenario allows up to $10$ time steps for our model to evolve we notice a great number and more varied results. In the first two rows the depicted digits bounce while moving towards one direction. In the third row the digit $0$ exhibits morphological changes and in the fourth row the digit $6$ gradually moves its closing intersection upwards to become a $0$. As our model is only trained with single frames of MNIST digits it is not constrained to show only movement or morphological changes. Rather it can vary both as seen in Figure~\ref{exp2_b}. The transmutation of the digit 6 to 0 is a probable, albeit unwanted, outcome under certain scenarios. In addition, we note that we are not providing any labels or additional information to the model during inference. In principle, one could condition the model to produce probable future frames by tuning the {\tt{choose}} procedure of Algorithm~\ref{algo1}. 

\subsection{Experiment 3: Realistic video data } 

\begin{figure*}[htb]
    \centering
    \subfloat[Moving MNIST
	\label{exp2_b}]{
        \includegraphics[height=4.7cm]{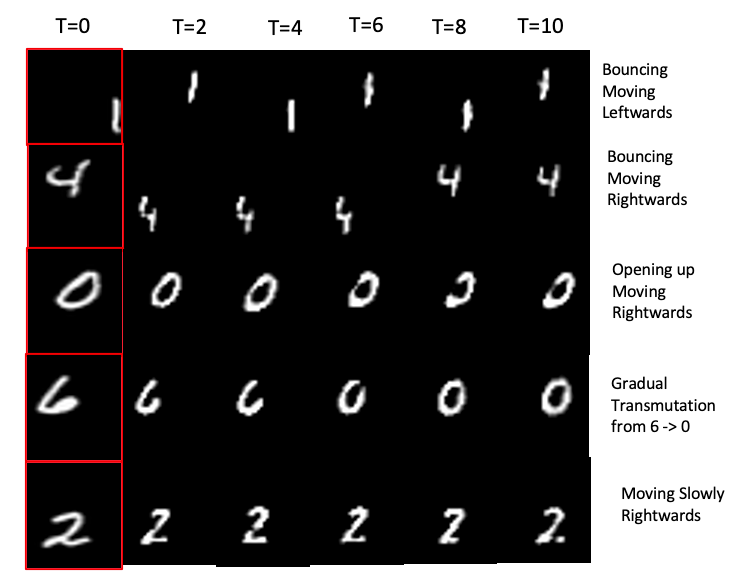}
        }
    \hfill
    \subfloat[KTH movement video sequences dataset
	\label{kth_exp}]{
        \centering
        \includegraphics[height=4.7cm]{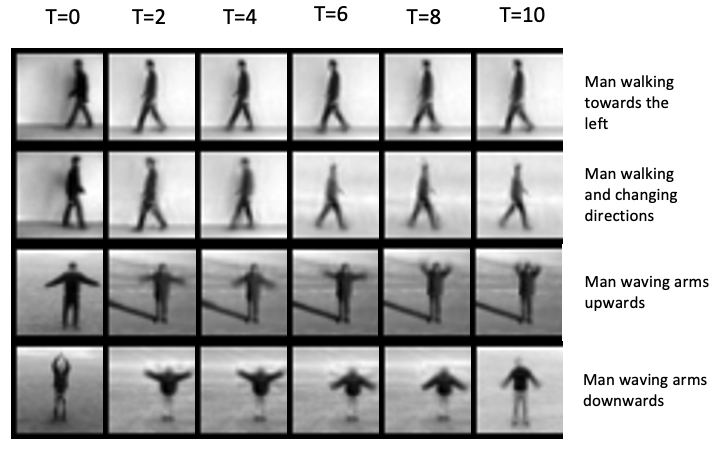}
    }
    \caption{Samples from Experiment 2 (a) and Experiment 3 (b). We are intersecting 5 cones trained on the moving MNIST dataset in (a) and the KTH movement video sequence dataset in (b). The latter is  representative for a real-world use-case scenario. Next to each row we added  an explanatory caption about the type of observed movement. Differences in image brightness in (b) are due to PyTorch's contrast normalization in the plotting function.}
    \label{fig:exp2}
\end{figure*}

As a final experimentation we use the KTH action dataset. Examples of the performance of the proposed algorithm are shown in Fig.~\ref{kth_exp}. Due to the computational constraints of the Poincar\'e VAE, which we are using as a base model, we are limited to one action at a time during training. We note how our algorithm retains characteristics like the shade of gray of the clothing while producing plausible futures. Each frame differs to the previous by 2 time instances giving ample time for the subject to change directions.  We believe that with a higher capacity network a similar performance can be achieved on more complex scenes and higher resolution videos. 

%real world performance is withing grasp of our algorithm. 

\iffalse
\begin{figure}
	\includegraphics[width=.6\linewidth]{images/exp2/exp2_a.png}
	\caption{Samples from Experiment 2. Intersecting 2 cones, on the left we exhibit vertical movements while on the right we exhibit horizontal movements. The arrows guide the direction of reading the figure. 
	\label{exp2_a}}
\end{figure}
\fi

	%\begin{figure}
%	\includegraphics[width=.5\linewidth]{images/exp2/exp2_b.png}
%	\caption{
%	\label{exp2_b}}
%	\end{figure}
	
\subsection{Discussion}
As the model is only trained as a VAE on single frames and not on sequences, the notion of time is not encoded in the weights of the network. Hence, all the resulting movement and predictive abilities are derived from our proposed algorithm and the natural embedding abilities of the Minkowski space-time.

We emphasize the time-agnostic nature of our algorithm. Our predictions are constrained in time but are probabilistic in nature. The proposed algorithm is able to produce multiple plausible futures. We believe this is a very important feature for future prediction and sequence extrapolation techniques as it can be used as an anomaly detection technique. Specifically, if one of the produced futures includes a hazardous situation, an automated system can adapt in order to avoid an outcome, enabling for example defensive driving capabilities in autonomous vehicles. 

Even though our method is in principle auto-regressive, it does not suffer from the accumulation of errors as it is both probabilistic and relies on efficient latent space sampling rather than the ability of a neural network to remember structural and temporal information about the image.

Furthermore, we believe that the quality of the predicted frames as well as the definition of the subspace from which the samples should be derived could be improved by incorporating the inferred thermodynamic entropy of the frame. We will explore the link between the information and thermodynamic entropy in future work. In addition, even though our framework is architecture agnostic, a customized architecture for the prediction task would be an intriguing direction. 

Finally, as our model allows us to find all probable scenarios that might exist, it can be used as a causal inference tool in the "potential outcomes" framework~\cite{Rubin2005}. Given a state we are able to probe possible scenarios and investigate plausible outcomes, hence, deduce causal relations within the data. In addition, by using the \emph{past} light cone $\mathcal{R^-}$, we are able to probe the events that could have led to an observed state enabling counterfactual analysis.

\section{Conclusion}
Machine learning techniques are able to build powerful representations of large quantities of data. We leverage this ability and propose that hard computer vision problems can be approached with minimal learning in an architecture agnostic manner. Strong mathematical and physical priors are the key. In this paper, we extend early Riemannian representation learning methods with the notion of Minkowski space-time as it is more suitable for causal inference. We further propose a novel algorithm to perform causally plausible image synthesis and future video frame prediction utilizing the special relativity concept of  light cones. We showed our algorithms' capabilities both in the synthetic Moving MNIST dataset as well as the real-world KTH dataset.

%\newpage
    \section{Broader Impact}
    In this paper we address an important theoretical topic. We believe that this work by itself does not have any ethical implications. We are proposing a generalized framework that can be used during causal reasoning applications and for future frame prediction. It is our conviction that applications using our framework that automate any use of equipment or provide medical diagnosis should have proper safety procedures in place to ensure the well-being of humans and animals. Our work incorporates insights from physics into the field of machine learning, allowing the development of a novel view on representation learning. This way, algorithms may gain insights about the world that are usually not consciously tractable by humans.
    There is no conflict of financial interests stemming from this work. 
  %  informed sampling 
   %  $t=2 : samples = 7741/100000$
	% $t=10 : samples = 56798/100000$
	 %$t=20 : samples = 70969/100000$
	% random sampling 
	% $t=2 : samples = 2/100000$
	% $t=10 : samples = 30876/100000$
	% $t=20 : samples = 71410/100000$
	 
	 \section{Acknowledgements}
	 The authors would like to thank L. Schmidtke and K. Kamnitsas for the fruitful discussions during the course of this project. Funding for this work was received by DoC of Imperial College London and the  MAVEHA (EP/S013687/1) grant (A.V., D.R., B.K.). H.B.R. received funding from the European Research Council (ERC) under the European Union’s Horizon 2020 research and innovation programme under grant agreement No 757646. The authors would also like to thank NVIDIA corp for the donation of GPU resources.
	 
	\bibliographystyle{splncs03}
	\bibliography{bibliography}

\begin{thebibliography}{10}
\providecommand{\url}[1]{\texttt{#1}}
\providecommand{\urlprefix}{URL }

\bibitem{arvanitidis2018latent}
Arvanitidis, G., Hansen, L.K., Hauberg, S.: Latent space oddity: on the
  curvature of deep generative models. In: International Conference on Learning
  Representations [2018]

\bibitem{Carroll1997}
Carroll, S.M.: {An introduction to general relativity: spacetime and geometry}
  [1997]

\bibitem{Ganea2018}
Ganea, O.E., B{\'{e}}cigneul, G., Hofmann, T.: {Hyperbolic Entailment Cones for
  Learning Hierarchical Embeddings}. ICML  [2018]

\bibitem{jayaraman2018timeagnostic}
Jayaraman, D., Ebert, F., Efros, A., Levine, S.: Time-agnostic prediction:
  Predicting predictable video frames. In: International Conference on Learning
  Representations [2019]

\bibitem{kosheleva2014observable}
Kosheleva, O., Kreinovich, V.: Observable causality implies lorentz group:
  alexandrov-zeeman-type theorem for space-time regions. Mathematical
  Structures and Modeling  [2014]

\bibitem{Kurutach2018}
Kurutach, T., Tamar, A., Yang, G., Russell, S., Abbeel, P.: {Learning plannable
  representations with causal infogan}. Advances in Neural Information
  Processing Systems  [2018]

\bibitem{LiuLLG18}
Liu, W., Luo, W., Lian, D., Gao, S.: {Future Frame Prediction for Anomaly
  Detection---A New Baseline}. In: {CVPR}. pp. 6536--6545. {IEEE} Computer
  Society [2018]

\bibitem{Mathieu2019}
Mathieu, E., Lan, C.L., Maddison, C.J., Tomioka, R., Teh, Y.W.: {Continuous
  Hierarchical Representations with Poincar\'e Variational Auto-Encoders}.
  NeurIPS  [2019]

\bibitem{MathieuCL15}
Mathieu, M., Couprie, C., LeCun, Y.: {Deep multi-scale video prediction beyond
  mean square error}. In: {ICLR} [2016]

\bibitem{Nickel2017}
Nickel, M., Kiela, D.: {Poincar{\'{e}} embeddings for learning hierarchical
  representations}. Advances in Neural Information Processing Systems [Nips]
  [2017]

\bibitem{Nickel2018}
Nickel, M., Kiela, D.: {Learning continuous hierarchies in the Lorentz model of
  hyperbolic geometry}. 35th International Conference on Machine Learning, ICML
  2018  [2018]

\bibitem{Pearl2019}
Pearl, J.: {The seven tools of causal inference, with reflections on machine
  learning}. Communications of the ACM  [2019]

\bibitem{pearl_glymour_jewell_2016}
Pearl, J., Glymour, M., Jewell, N.P.: Causal inference in statistics a primer.
  Wiley [2016]

\bibitem{Peters2019}
Peters, J., Janzing, D., Sch{\"{o}}lkopf, B.: {Elements of Causal Inference :
  Foundations and Learning Algorithms}. The MIT Press [2019]

\bibitem{Rubin2005}
Rubin, D.B.: {Causal inference using potential outcomes: Design, modeling,
  decisions}. Journal of the American Statistical Association  100 [2005]

\bibitem{schuldt2004recognizing}
Schuldt, C., Laptev, I., Caputo, B.: Recognizing human actions: a local svm
  approach. In: Proceedings of the 17th International Conference on Pattern
  Recognition, 2004. ICPR 2004. vol.~3, pp. 32--36. IEEE [2004]

\bibitem{DBLP:journals/corr/SrivastavaMS15}
Srivastava, N., Mansimov, E., Salakhutdinov, R.: Unsupervised learning of video
  representations using lstms. ICML  [2015]

\bibitem{NIPS2015_5971}
Sun, K., Wang, J., Kalousis, A., Marchand-Maillet, S.: Space-time local
  embeddings. In: NIPS [2015]

\bibitem{Tulyakov0YK18}
Tulyakov, S., Liu, M., Yang, X., Kautz, J.: {MoCoGAN: Decomposing Motion and
  Content for Video Generation}. In: {CVPR}. pp. 1526--1535. {IEEE} Computer
  Society [2018]

\bibitem{VillegasYHLL17}
Villegas, R., Yang, J., Hong, S., Lin, X., Lee, H.: {Decomposing Motion and
  Content for Natural Video Sequence Prediction}. In: {ICLR} [2017]

\bibitem{DBLP:journals/corr/VillegasYZSLL17}
Villegas, R., Yang, J., Zou, Y., Sohn, S., Lin, X., Lee, H.: Learning to
  generate long-term future via hierarchical prediction. ICML  [2017]

\bibitem{cvpr/VondrickPT16}
Vondrick, C., Pirsiavash, H., Torralba, A.: {Anticipating Visual
  Representations from Unlabeled Video}. In: {CVPR}. pp. 98--106. {IEEE}
  Computer Society [2016]

\bibitem{nips/VondrickPT16}
Vondrick, C., Pirsiavash, H., Torralba, A.: {Generating Videos with Scene
  Dynamics}. In: {NIPS}. pp. 613--621 [2016]

\bibitem{Zeeman1964}
Zeeman, E.C.: {Causality implies the Lorentz group}. Journal of Mathematical
  Physics  5[4],  490--493 [1964]

\end{thebibliography}

\end{document}

\maketitle
	
	\section{Step-by-Step Visualization of the Proposed Algorithm}
	In order help the reader further understand the intuition behind our proposed algorithm we will be conducting a mental experiment with the help of some visual queues. Note the figures will follow the convention of 2+1 dimensional euclidean space, simply for ease of understanding. 
	
		\begin{figure}[ht]
	\includegraphics[width=.6\linewidth]{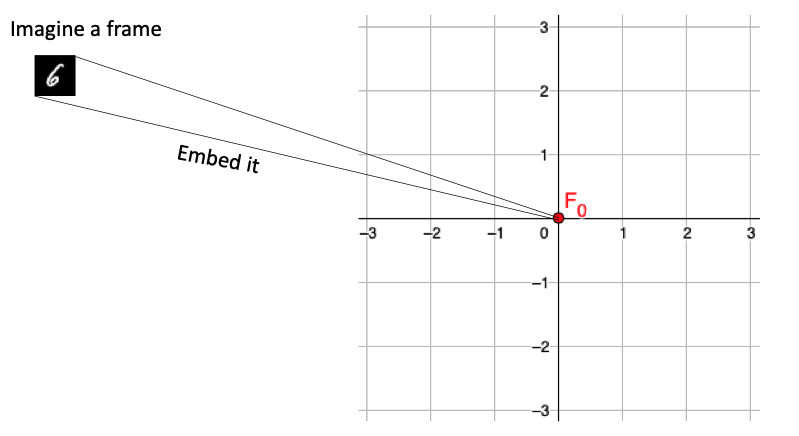}
	\caption{Lets assume a frame $F_0$ that we embed in our space
	\label{fig1}}
	\end{figure}
		\begin{figure}[ht]
	\includegraphics[width=.6\linewidth]{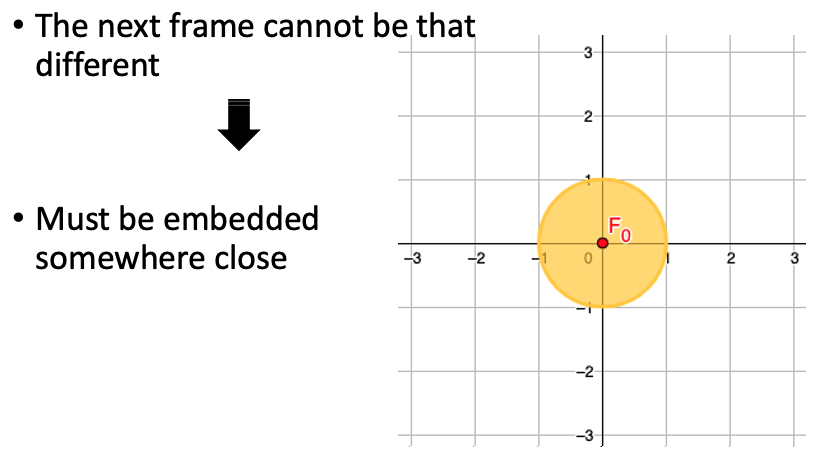}
	\caption{Where can the second frame lie ? 
	\label{fig1}}
	\end{figure}
	
		\begin{figure}[ht]
	\includegraphics[width=.8\linewidth]{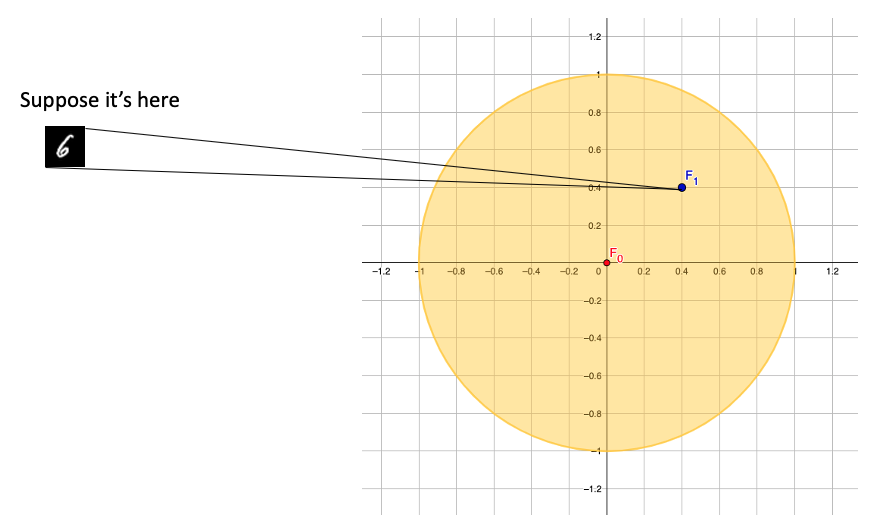}
	\caption{Embed the second frame 
	\label{fig1}}
	\end{figure}
	
	\begin{figure}[ht]
	\includegraphics[width=.9\linewidth]{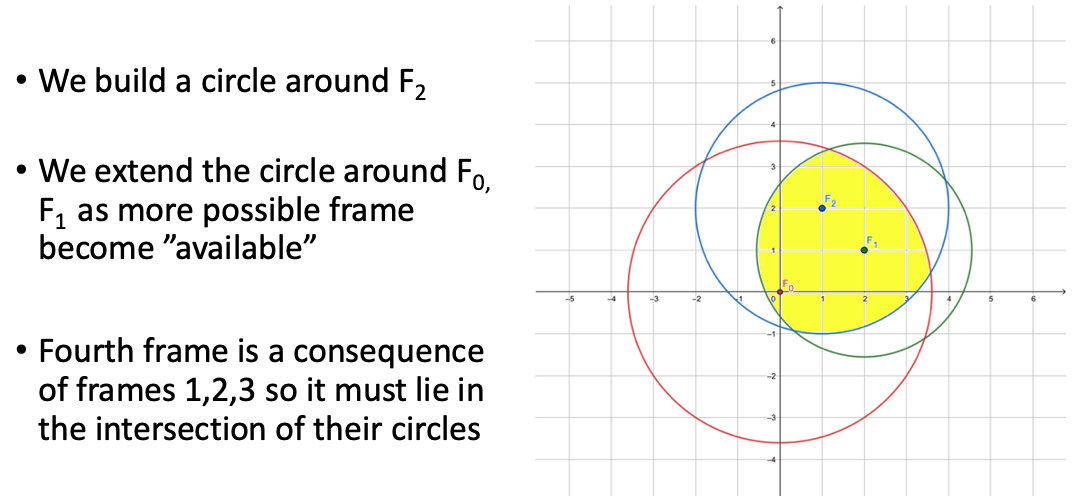}
	\caption{Embed a third frame $F_2$ and increase the circles of frames 0,1 the fourth frame has to lie in the intersection of the circles.
	\label{fig1}}
	\end{figure}

	\begin{figure}[ht]
	\includegraphics[width=.9\linewidth]{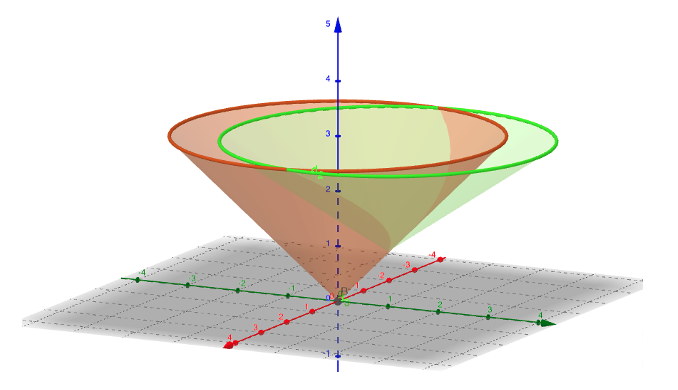}
	\caption{In space-time this would look like the intersection of cones. Due to the fact that the increasing radius of the 2D circles create a cone in 3D.
	\label{fig1}}
	\end{figure}
	\begin{figure}[ht]
	\includegraphics[width=.9\linewidth]{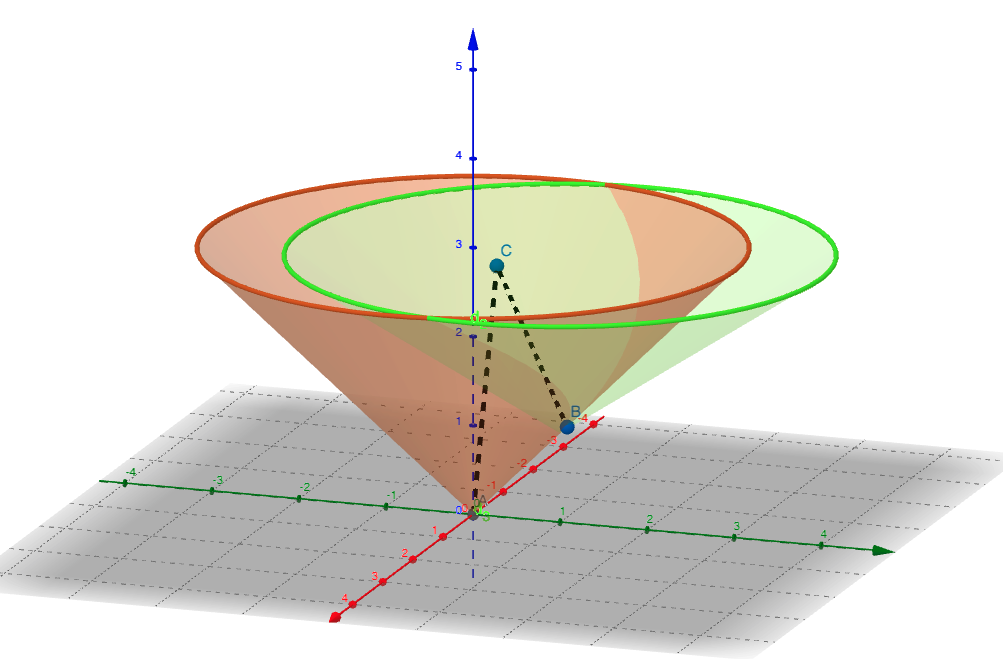}
	\caption{Two potential causal paths from points a,b to a new point c. Note that if its these points represent a sequence then the causal path will have to pass from $A\rightarrow B\rightarrow C$
	\label{fig1}}
	\end{figure}
	
	\newpage
	
	\section{Poincar\'e Ball} 
	
	As stated in the main paper Riemannian Geometry can be seen as a curved generalization of Euclidean space. In this section we will be focusing on the Poincar\'e ball Riemannian manifold, as it forms the basis of our implementation. We note that there is no theoretical reason why we extend the Poincar\'e VAE~\cite{Mathieu2019} other than simplicity of implementation and proven results in the image domain.
	
Many works rely on a Poincar\'e ball, as has been argued \cite{Nickel2017,Mathieu2019,Ganea2018} that embedding the latent space on a Poincar\'e Ball, that is, a hyperbolic space with negative curvature, allows one to naturally embed continuous hierarchical relationships between data points. This follows from the qualitative properties of such a hyperbolic space:
    \begin{enumerate}
        \item The entirety of the Poincar\'e Ball $\mathcal{B}_c^d$ is contained within a hypersphere of radius $1/\sqrt{c}$ and dimensionality $d$, in what amounts to \textit{compactification} of infinite space.
        \item The distance function (and thus area element) of this space grows rapidly as one approaches the edges of $\mathcal{B}_c^d$, such that reaching the edge would require traversing an infinite distance in latent space.
        \item This behaviour naturally emulates the properties of hierarchical trees, whose size grows exponentially as new branches "grow" from previously existing branches.
        \end{enumerate}
    Quantitatively, the space $\mathcal{B}_c^d$ is endowed with a metric tensor $g^c$ which relates to flat Euclidean space as follows
    \begin{equation} \label{poinceucl}
        g^c(\bold{r})=\left(\frac{2}{1-c |\bold{r}|^2}\right)^2 g_e(\bold{r})
    \end{equation}
    where \textbf{r} is a $d$-dimensional vector in latent space and $g_e(\bold{r}$ the Euclidean metric. As a result, the distance element in $\mathcal{B}_c^d$ may be written, in spherical coordinates,
    \begin{equation}
        ds^2=\left(\frac{2}{1-c r^2}\right)^2 (dr^2+r^2d\Omega^2_d)
    \end{equation}
    where $r=|\bold{r}|$ is the radius from the origin of the space and $d\Omega_d$ is the differential solid angle element in $d$ dimensions. As explained qualitatively above, it easy to see that the distance element diverges as $r\rightarrow 1/\sqrt{c}$, thus encoding the infinite hypervolume contained near the edges of the Poincar\'e Ball. Furthermore, as $c\rightarrow 0$, the radius of the Poincar\'e Ball becomes infinity and $g_c(\bold{z})\rightarrow g_e(\bold{z})$, up to a constant rescaling of the coordinates.
    Let $\gamma:t\rightarrow\gamma(t)$ be a curve in $\mathcal{B}_c^d$, where $t\in[0,1]$ such that its length is defined by 
    \begin{equation}
        L(\gamma(t))=\int_0^1\sqrt{ds^2(t)}dt=\int_0^1\sqrt{\bold{v}^T(t)\, \bold{\hat g^c}\,\bold{v}(t)}\,dt
    \end{equation}
    where $\bold{\hat g^c}$ is the matrix form of $g^c$ and $\bold{v}\equiv\frac{d\bold{r}}{dt}$ is the trajectory's velocity vector. In component form, this reads:
    \begin{equation}\label{distfull}
    =\int_0^1\sqrt{\sum_{\mu=1}^d\sum_{\nu=1}^d\frac{dx^\mu(t)}{dt}g_{\mu\nu}^c\frac{dx^\nu(t)}{dt}\,dt}=\int_0^1 \left(\frac{2}{1-c r^2(t)}\right)^2\,\sqrt{\bold{v}^T\bold{v}}\,dt
    \end{equation}
    where $x^\mu$ represents each coordinate, $g^c_{\mu\nu}$ is the component form of $\bold{\hat g^c}$ and in the last step we have used (\ref{poinceucl}). In hyperbolic space, "straight lines" are defined by \textit{geodesics} $\gamma_g(t)$, that is, curves of constant speed and least distance between points $\bold{x}$ and $\bold{y}$. Thus,
    \begin{equation} \label{geo}
    \gamma_g(t)=\text{argmin}\left[L(\gamma(t))\right]_{\gamma(0)=\bold{x}}^{\gamma(1)=\bold{y}}\quad\text{and}\quad \left|\frac{d\gamma(t)}{dt}\right|=1 .
    \end{equation}
    With (\ref{distfull}) and (\ref{geo}), one may show that the distance function $d^c(\bold{x},\bold{y})$ between two points $\bold{x}$ and $\bold{y}$ on $\mathcal{B}_c^d$ can be computed to yield:
    \begin{equation} \label{dist}
    d^c(\bold{x},\bold{y})=\frac{1}{\sqrt{c}}\text{arccosh}\left( 1+2c\frac{|\bold{x}-\bold{y}|^2}{(1-c|\bold{x}|^2)(1-c|\bold{y}|^2)}\right)
    \end{equation}
    
\section{Wrapped Normal}
While embedding data on a Riemannian space with the use of a Riemannian VAE, it is important to embed the used distribution in the space as well. Multiple ways have been proposed to perform this process. We will be following the wrapped normal distribution~\cite{grattarola_livi_alippi_2019,Mathieu2019}.

In this approach a normal distribution is mapped onto the manifold using the manifold's exponential map. Given a normal distribution $z_e\sim\mathcal{N}(0,\Sigma)$ then the Riemannian sample $z = exp^c_\mu(\frac{z_e}{\lambda^c_\mu})$.  The distribution's density can be described as:
\begin{equation}
    \mathcal{N}^W_{\mathcal{B}^d_c}(z|\mu,\Sigma)= \frac{dv^W(z|\mu,\Sigma)}{d\mathcal{M} (z)} = \mathcal{N}( \lambda^c_\mu log_\mu(z)|0,\Sigma) (\frac{\sqrt{c} d^c_p(\mu,z)}{sinh(\sqrt{c}d^c_p(\mu,z))})^{d-1}
\end{equation}
As $c\rightarrow0$ the Euclidean normal distribution can be obtained.

\section{Architectural Considerations}
In order to properly embed information on a manifold a set of considerations have to be taken into account as developed by Ganea et al.~\cite{Ganea2018networks}. In this paper we are following the architectural guidance of \cite{Ganea2018networks,Mathieu2019} about the last layer of the encoder and the first of the decoder. Specifically in the encoder we use the frechet mean as calculated by the exponential mapping $exp^c_0$ and a solftplus variance $\sigma$. In terms of the decoder we utilize the gyroplane layer as developed by \cite{Ganea2018networks,Mathieu2019}. Our architecture follows the consideration from \cite{Mathieu2019} with the additions of the extended mapping using equations 5,6 from the main paper and increased capacity of the hidden layers as our input is $32\times32$ rather than the original $28\times28$. 

\paragraph{Optimization:} In terms of optimization we tested both a Riemannian stochastic gradient descent as seen in \cite{Nickel2017} and \cite{Mathieu2019}'s approach of using the exponential mapping to bring the model's parameters onto the manifold. As there was inconsequential practical differences between the two approaches and as both are theoretically sound we opted with \cite{Mathieu2019}'s approach for computational simplicity. 

\section{Online learning and Anomaly Detection}
A limitation of causal analysis is that it fails to include unseen causal sources. For example in the case of an autonomous vehicle simulation the addition of a second car by the researcher would entail a causal anomaly in the world model of the autonomous vehicle where it was the only vehicle in existence. It is obvious that causal future predictions like these are impossible to perform as the system is assumed to be impervious to modifications from outside sources. 

However, by presenting an anomaly like that and mapping it on our world model we are able to raise an anomaly flag. If an observation is made that falls outside the perceived light cone of the system then if an event like this has happened before we are able to adapt the aperture of the cone, or as in the case of the scientist inserting a new vehicle, re-structure the world model of the system based on the new observation. This can be considered as a method of online fine-tuning. We believe that beyond fully retraining our model we are able to adjust the embedding space and subsequently the light cones by modifying the metric with the use of free parameters. Investigations on which method is optimal will follow in future work.

\section{Code}
The proposed algorithm was implemented in PyTorch and the code will be made available by the time of the conference.
	\bibliographystyle{splncs03}
	\bibliography{bibliography}